%% file: main.tex
\documentclass[10pt,twocolumn,letterpaper]{article}

\usepackage{cvpr}              
\usepackage{graphicx}	
\usepackage{amsmath}	
\usepackage{amssymb}	
\usepackage{booktabs}
\usepackage{times}
\usepackage{microtype}
\usepackage{epsfig}
\usepackage[table,xcdraw,dvipsnames]{xcolor}
\usepackage{caption}
\usepackage{float}
\usepackage{placeins}
\usepackage{color, colortbl}
\usepackage{stfloats}
\usepackage{enumitem}
\usepackage{tabularx}
\usepackage{xstring}
\usepackage{multirow}
\usepackage{multicol}
\usepackage{xspace}
\usepackage{url}
\usepackage{subcaption}
\usepackage{algcompatible}
\usepackage{algorithm}

\input{preamble}
\definecolor{cvprblue}{rgb}{0.21,0.49,0.74}
\usepackage[pagebackref,breaklinks,colorlinks,allcolors=cvprblue]{hyperref}

\def\paperID{***} 
\def\confName{CVPR}
\def\confYear{2026}

\title{DYMAPIA: A Multi-Domain Framework for Detecting AI-based Video Manipulation}

\author{Md Shohel Rana\\
School of Computing\\
Georgia Southern University\\
Statesboro, GA 30460, USA\\
{\tt\small mrana@georgiasouthern.edu}
\and
Andrew H. Sung\\
School of Computing Sciences \& Computer Engineering\\
The University of Southern Mississippi\\
Hattiesburg, MS 39406, USA\\
{\tt\small andrew.sung@usm.edu}
}

\begin{document}
\maketitle
\input{sec/0_abstract}    
\input{sec/1_intro}
\input{sec/2_related}
\input{sec/4_method}
\input{sec/3_hypothesis}
\input{sec/5_conclusion}

{
    \small
    \bibliographystyle{ieeenat_fullname}
    \bibliography{main}
}


\end{document}

%% file: sec/0_abstract.tex
\begin{abstract}
AI-generated media are advancing rapidly, raising pressing concerns for content authenticity and digital trust. We introduce DYMAPIA, a multi-domain Deepfake detection framework that fuses spatial, spectral, and temporal cues to capture subtle traces of manipulation in visual data. The system builds dynamic anomaly masks by combining evidence from Fourier spectra, local texture descriptors, edge irregularities, and optical flow consistency, which highlight tampered regions with fine spatial accuracy. These masks guide DistXCNet, a lightweight classifier distilled from Xception and optimized with depthwise separable convolutions for fast, region-focused classification. This joint design achieves state-of-the-art results, with accuracy and F1-scores exceeding 99\% on FF++, Celeb-DF, and VDFD benchmarks, while keeping the model compact enough for real-time use. Beyond outperforming existing full-frame and multidomain detectors, DYMAPIA demonstrates deployment readiness for time-critical forensic tasks, including media verification, misinformation defense, and secure content filtering.
\end{abstract}

%% file: sec/1_intro.tex
\section{Introduction}
\label{sec:intro}
The rapid progress of generative models \cite{face2face, CycleGAN, Synthesizing_Obama}, particularly Generative Adversarial Networks (GANs) \cite{gans} and diffusion-based synthesis, has made it possible to generate photorealistic images and videos that are often indistinguishable from authentic content. These so-called Deepfakes are no longer confined to research labs; they are widely accessible through open source tools and user-friendly platforms. As a result, manipulated media have already been exploited in disinformation campaigns, identity fraud, and even falsified evidence, posing serious threats to journalism, digital forensics, and national security \cite{patrini2018state, ajder2019deepfake}. In this context, ensuring the authenticity of visual media has become a pressing research challenge with direct societal implications.

Over the last few years, many detection strategies have been proposed, ranging from convolutional neural networks (CNNs) to attention-based models and frequency-domain filters. While these approaches have advanced the field, two common limitations remain. First, most methods rely heavily on single-domain forensic cues, such as pixel-level irregularities or frequency distortions. These cues can be easily suppressed by advanced synthesis pipelines that are deliberately designed to hide such artifacts. Second, detection systems often process full frames of images or videos. This not only adds significant computational overhead but also dilutes the forensic signal with irrelevant background information, making detection less reliable in unconstrained or long-form media.

To address these gaps, we introduce DYMAPIA (Dynamic Masking and Pixel-wise Inconsistency Analysis), a unified multi-domain framework for detecting manipulations in both images and videos. Instead of depending on a single forensic signal, DYMAPIA fuses complementary cues from three perspectives:

\begin{itemize}
    \item \textbf{Frequency domain analysis} to capture spectral distortions that arise from frame interpolation and content blending \cite{Style_Based, Deep_video_portraits}.
    \item \textbf{Texture, edge, and contour analysis} using Local Binary Patterns (LBP) \cite{lbp}, which reveal subtle inconsistencies in skin details, facial boundaries, and background textures.
    \item \textbf{Temporal consistency evaluation} with dense optical flow \cite{deepfakes_creation_detection}, which exposes abrupt transitions or implausible motion patterns across consecutive frames.
\end{itemize}

Each of these analyses generates an anomaly mask, and the masks are merged into a dynamic composite mask that highlights manipulated regions at pixel-level granularity. This approach improves interpretability by showing where a manipulation occurs, while also strengthening robustness by combining cues that are difficult to suppress simultaneously.

For classification, we propose \textbf{DistXCNet}, a compact convolutional architecture distilled from XceptionNet \cite{xceptionnet}. Unlike conventional detectors that operate on entire frames, DistXCNet learns directly from the dynamic masks, focusing on regions most likely to contain forensic evidence. Its design is based on depthwise separable convolutions, which drastically reduce the parameter count (fewer than 14K) while preserving discriminative power across multiple scales. This allows the model to run efficiently on resource-limited hardware, enabling real-time deployment.

The \textbf{DYMAPIA} is designed with practical deployment in mind. We conduct runtime and memory profiling to evaluate its readiness for applications such as forensic triage, misinformation suppression, and secure content verification. Unlike most prior work, which targets only video detection, our framework generalizes naturally to manipulated images, expanding its usefulness across a broader range of multimedia forensics tasks.

Our main contributions can be summarized as follows:
\begin{itemize}
    \item \textbf{A unified detection framework (DYMAPIA)}: We combine spectral, spatial, temporal cues into anomaly masks that pinpoint manipulated regions.

    \item \textbf{A lightweight classifier (DistXCNet)}: We introduce a distilled Xception-based CNN that operates on sparse masks, achieving strong accuracy with $<$14K parameters.

    \item \textbf{Comprehensive validation}: We evaluate our approach on FF++, Celeb-DF, and VDFD datasets, showing consistent improvements in accuracy, F1-score, and generalization over state-of-the-art baselines.

    \item \textbf{Deployment-oriented analysis}: We profile runtime, memory usage, and resilience under compression and low-motion conditions, demonstrating real-world viability for forensic and security applications.
\end{itemize}

Our experiments confirm that DYMAPIA significantly improves detection accuracy and interpretability while remaining lightweight and fast enough for deployment. Beyond cybersecurity, the framework has implications for journalism, law enforcement, and digital forensics, where trustworthy media verification is critical to countering the risks of AI-generated disinformation.

The rest of this paper is organized as follows: Section~\ref{sec:related} reviews prior Deepfake detection methods; Section~\ref{sec:method} details the DYMAPIA framework; Section~\ref{sec:hypothesis} presents experimental validation and performance analysis; and Section~\ref{sec:conclusion} concludes with key findings and directions for future work.


%% file: sec/2_related.tex
\section{Related Work}
\label{sec:related}
The growing sophistication of Deepfake technology has raised serious concerns about the integrity and reliability of digital media. As generation techniques advance, detection research has evolved from early heuristic approaches to modern deep learning–driven methods. In this section, we review prior work across five categories: handcrafted feature–based methods, deep learning models, multi-domain strategies, limitations of existing frameworks, and adaptive approaches. Together, these studies highlight the research gaps that motivate the design of DYMAPIA.

\subsection{Early Detection Methods Based on Handcrafted Features}
The first generation of Deepfake detection relied heavily on handcrafted features, often focusing on physiological or geometric inconsistencies. For example, eye-blinking irregularities and facial landmark distortions were among the earliest cues used to separate authentic and manipulated videos \cite{eye_blink}. While these techniques were initially effective, they quickly became obsolete as generative methods improved. Advanced frameworks such as DeepFaceLab and the First Order Motion Model (FOMM) \cite{DeepFaceLab} introduced realistic facial reenactment and lip synchronization, masking the kinds of irregularities that handcrafted detectors relied on. Neural rendering approaches, such as Neural Voice Puppetry \cite{Neural_Voice_Puppetry}, further increased realism, making purely heuristic detection infeasible.

\subsection{Deep Learning (DL)-Based Detection Models}
With the limitations of handcrafted features, deep learning emerged as a dominant direction. Afchar et al. introduced MesoNet \cite{MesoNet}, one of the first CNN-based models purpose-built for Deepfake detection, while Rossler et al. created FaceForensics++ (FF++) \cite{ffpp}, which remains a cornerstone dataset for benchmarking. These methods demonstrated the potential of convolutional architectures to capture subtle synthetic artifacts.

At the same time, researchers have emphasized that deep learning is not a complete solution. Several studies \cite{Rana_SLR, reality_under_threat, rana_tutorial} point out that classical machine learning techniques, when paired with carefully engineered features, can still outperform deep networks in specific settings. Such methods offer interpretability and lower computational cost, leading to growing interest in hybrid approaches that combine the strengths of both machine learning (ML) and deep learning (DL) \cite{Rana_ML, advanced_deepfake}.

\subsection{Multi-Domain Approaches in Deepfake Detection}
To improve robustness, researchers have increasingly turned to multi-domain feature analysis. Frequency-domain methods, such as those by Durall et al. \cite{Unmasking_DeepFakes}, exploit spectral distortions introduced during synthesis. Texture-based detectors, including Li et al.’s Face X-ray model \cite{x_ray}, focus on inconsistencies in skin patterns and blending boundaries.

Temporal analysis has also proven valuable. Optical flow–based approaches, such as those by Güera and Delp \cite{rnn_guera}, identify unnatural frame-to-frame transitions, while Nguyen et al. \cite{deepfakes_creation_detection} apply spatio-temporal modeling to reveal subtle inconsistencies in motion.

Survey work has reinforced the importance of such integrated designs. Ahmed et al. \cite{visual_deepfake} categorize detection methods into spatial, temporal, frequency, and spatio-temporal domains, underlining the advantages of fusing complementary cues. More recent multimodal frameworks go a step further by combining audio and visual information, for instance in MultiModalTrace \cite{Multimodaltrace} and Person-of-Interest detection \cite{Person_of_Interest}, demonstrating the potential of cross-modal learning for resilience against advanced Deepfakes.

\subsection{Limitations of Existing Approaches and Motivation for DYMAPIA}
Despite these advances, three persistent challenges remain. First, many detection models are single-domain, making them brittle against adaptive attacks that conceal or minimize specific artifacts. Second, dataset diversity is limited, with most benchmarks focused on a narrow set of manipulations, leading to poor generalization in the wild \cite{dfdc}. Third, computational efficiency is often overlooked, some models achieve high accuracy but require heavy architectures that are impractical for real-time deployment \cite{Balafrej2024Enhancing}.

DYMAPIA is designed to address these gaps. By fusing spectral, spatial, and temporal cues, the framework strengthens robustness against diverse manipulations. Its validation extends beyond video to also include still images, broadening applicability. Finally, we explicitly benchmark runtime and memory performance to demonstrate its practicality for real-world forensic and security contexts.

\subsection{Security Implications \& Adaptive Detection Strategies}
The significance of Deepfake detection extends beyond technical performance. In practice, robust detectors play a vital role in preventing misinformation, supporting forensic investigations, and protecting public trust. Chesney and Citron \cite{Chesney2019DeepFakes} highlight the risks of synthetic media in political manipulation and fraud, while Verdoliva \cite{Media_Forensics} emphasizes the broader cybersecurity implications.

An emerging line of research focuses on adaptive frameworks that evolve alongside generative models. Mirsky and Lee \cite{Mirsky} discuss adversarial countermeasures and the need for detection systems that can adapt as new attack methods appear. DYMAPIA follows this direction by implementing dynamic anomaly masking and systematic failure analysis, ensuring transparency, adaptability, and resilience as Deepfake techniques continue to advance.

%% file: sec/4_method.tex
\section{Methodology}
\label{sec:method}
\begin{figure}[!ht]
  \centering
  \includegraphics[width=1.0\linewidth]{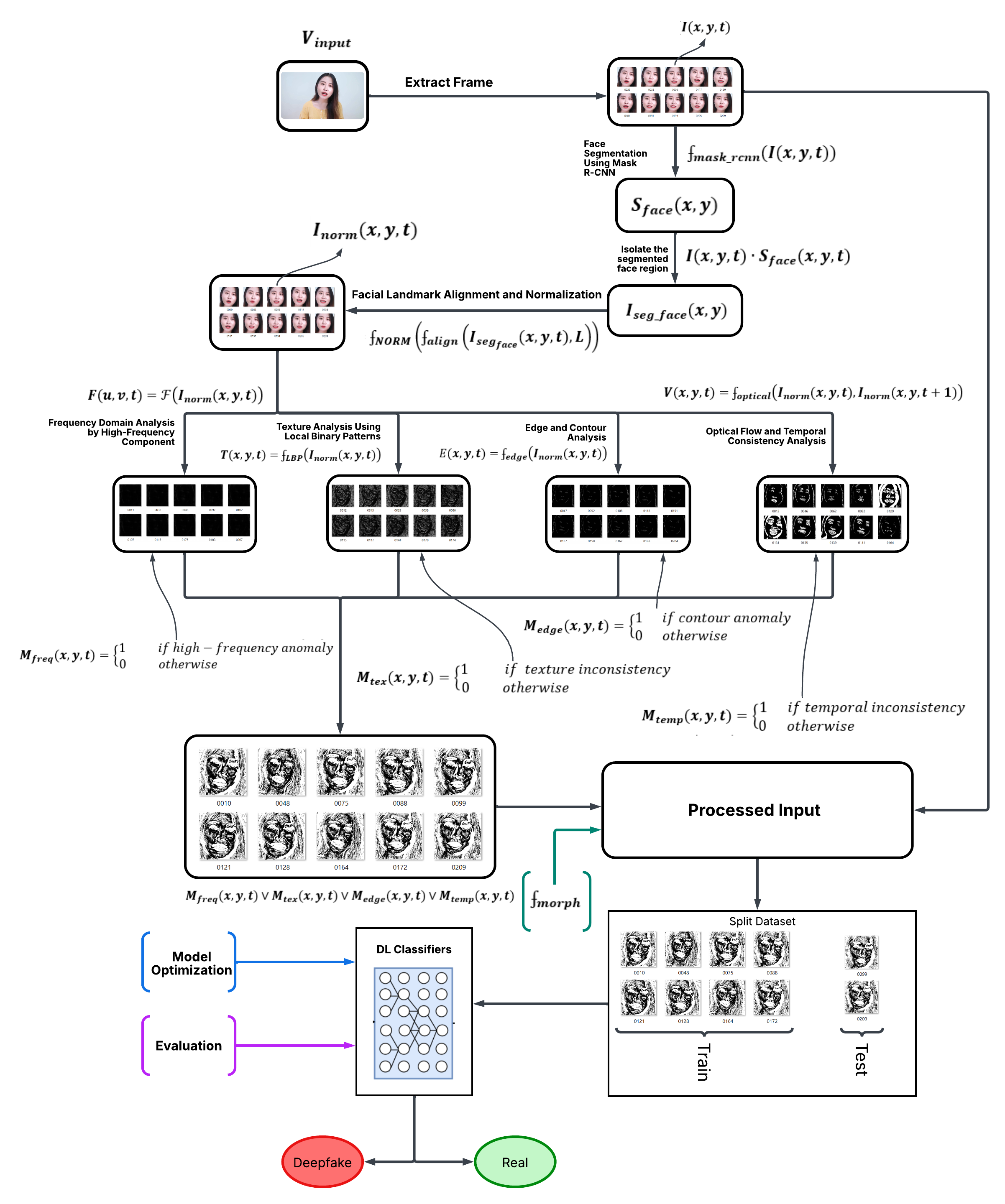}
  \caption{A pipeline of the proposed DyMAPIA approach.}
  \label{fig:pipeline_of_DyMapia}
\end{figure}

The proposed \textbf{DYMAPIA} framework introduces a multistage methodology for Deepfake detection that combines anomaly localization with efficient classification. At its core, DYMAPIA generates dynamic masks that expose pixel-level inconsistencies in manipulated faces and then leverages these masks to guide a lightweight deep learning model. The methodology unfolds in two main phases: (1) dynamic mask generation to identify potential artifacts and (2) mask-guided classification to determine whether the content is authentic or manipulated. The full pipeline is shown in Figure~\ref{fig:pipeline_of_DyMapia} and is formally described in Algorithm~\ref{alg:dymapia}.

\subsection{Dynamic Mask Generation for Manipulation Detection}
\begin{figure}[!ht]
    \centering
    \includegraphics[width=0.75\linewidth]{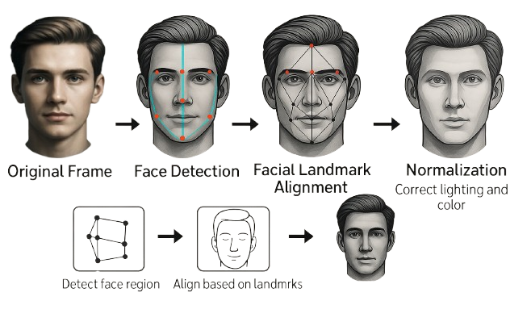}
    \caption{Preprocessing pipeline including face detection, alignment, and normalization.}
    \label{fig:step1}
\end{figure}

The first stage of DYMAPIA focuses on localizing manipulation artifacts through a sequence of anomaly detection steps. As shown in Figure~\ref{fig:step1}, the pipeline begins with preprocessing, where video frames undergo face detection, alignment, and normalization. To achieve precise localization, DYMAPIA employs Mask R-CNN~\cite{he2018maskrcnn}, a segmentation model capable of extracting facial components such as the eyes, nose, and mouth. This step reduces background noise and allows subsequent analyses to focus only on areas most susceptible to Deepfake artifacts.

While Mask R-CNN provides high accuracy, its computational cost may hinder scalability. In future work, more efficient alternatives (e.g., HRNet or MobileNet) could be adopted for large-scale or real-time deployment. Another consideration is that this segmentation is optimized for uncompressed data; compression artifacts introduced in MP4 or other formats may obscure anomalies and reduce segmentation quality. 

\begin{figure}[!ht]
    \centering
    \includegraphics[width=0.75\linewidth]{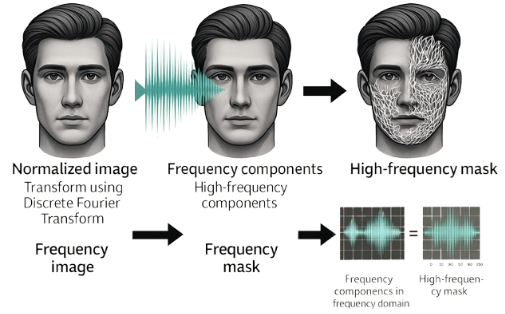}
    \caption{Frequency Domain Analysis using DFT and anomaly mask generation.}
    \label{fig:step3}
\end{figure}

Following preprocessing, the framework proceeds through a sequence of complementary analysis steps, each designed to capture unique indicators of synthetic manipulation. 

\textbf{Frequency Domain Analysis.} After preprocessing, DYMAPIA applies the Discrete Fourier Transform (DFT) to project the normalized face into the frequency domain (Figure~\ref{fig:step3}). Deepfake synthesis and compression often leave behind high-frequency spectral artifacts that are not easily visible in the spatial domain. By isolating these frequency components, the system generates a frequency anomaly mask that highlights subtle distortions in textures, lighting transitions, and edges.

\begin{figure}[ht]
    \centering
    \includegraphics[width=0.8\linewidth]{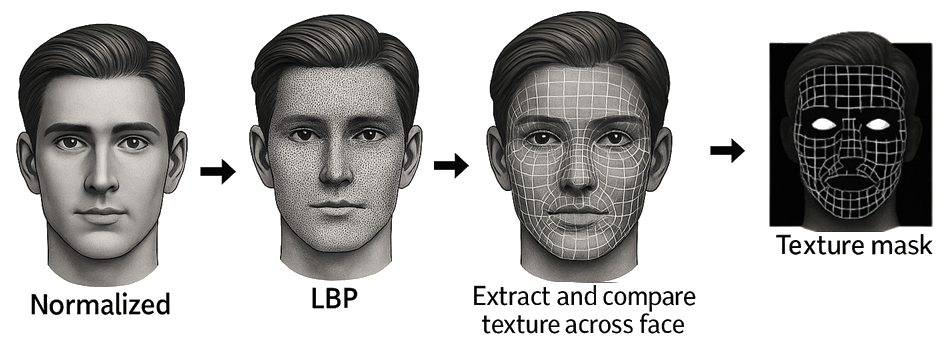}
    \caption{Texture Analysis using LBP to highlight inconsistencies.}
    \label{fig:step4}
\end{figure}

\textbf{Texture Analysis.} The next step leverages Local Binary Patterns (LBP)~\cite{lbp} to detect inconsistencies in the surface texture of the face (Figure~\ref{fig:step4}). Deepfake algorithms frequently fail to reproduce realistic skin granularity, instead introducing artifacts such as blurring, patch repetition, or irregular shading. LBP encodes local intensity variations into binary values, which are compared against expected distributions to produce a texture inconsistency mask.

\begin{figure}[ht]
    \centering
    \includegraphics[width=0.7\linewidth]{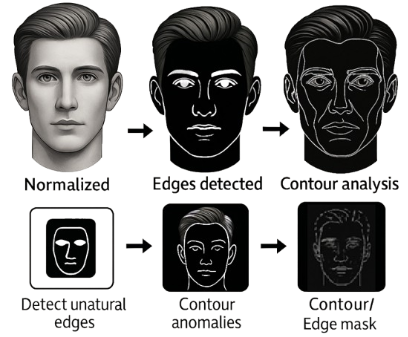}
    \caption{Edge and contour detection to capture unnatural transitions.}
    \label{fig:step5}
\end{figure}

\textbf{Edge and Contour Detection.} Structural abnormalities are captured in the third stage using edge detection techniques, such as the Canny operator (Figure~\ref{fig:step5}). Deepfake manipulations often leave unnatural or sharp transitions around facial features like the jawline or mouth. These irregularities are preserved in an edge anomaly mask, which exposes synthetic blending regions and misplaced contours.

\begin{figure}[ht]
    \centering
    \includegraphics[width=0.7\linewidth]{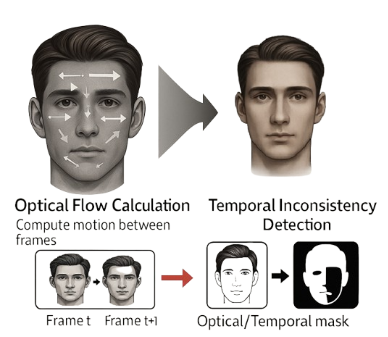}
    \caption{Temporal consistency analysis via optical flow.}
    \label{fig:step6}
\end{figure}

\textbf{Optical Flow and Temporal Consistency Analysis.} For video sequences, DYMAPIA also analyzes motion coherence across consecutive frames (Figure~\ref{fig:step6}). Using dense optical flow, the system evaluates whether motion patterns are temporally consistent. Deepfake videos often contain temporal anomalies such as inconsistent blinking or desynchronized lip movements. The optical flow results are used to generate a temporal inconsistency mask that flags regions exhibiting unrealistic dynamics.

\textbf{Mask Integration and Refinement.} All anomaly masks such as frequency, texture, edge, and temporal are combined using a logical OR operation to create a composite dynamic mask. This ensures that any region flagged by at least one modality is retained. Since each analysis contributes unique forensic evidence, their integration provides a more complete view: frequency masks capture spectral distortions, texture masks highlight skin-level irregularities, edge masks reveal boundary artifacts, and temporal masks expose motion-based inconsistencies. At the final step, the composite mask is refined through morphological operations (e.g., erosion and dilation) to reduce isolated noise and improve spatial continuity. The refined mask is then forwarded to the classification stage, where only the highlighted regions are analyzed, improving both efficiency and interpretability.

\subsection{Proposed DistXCNet Model}
The second stage of DYMAPIA employs DistXCNet, a lightweight convolutional neural network designed specifically for mask-guided classification. The key idea is that, rather than analyzing an entire RGB frame, the classifier operates only on refined anomaly masks produced by DYMAPIA. This approach achieves three goals simultaneously: it reduces redundant background computation, increases interpretability by focusing on manipulation-prone regions, and lowers the computational footprint without compromising accuracy.

\textbf{Input Representation.} Given a normalized grayscale frame $I(x, y, t)$ at time $t$ and its refined anomaly mask $M_{\text{refined}}(x, y, t) \in {0,1}$, the masked input is defined as:

\begin{equation}
I_{\text{masked}}(x, y, t) = I(x, y, t) \cdot M_{\text{refined}}(x, y, t)
\end{equation}

This ensures that only pixels within anomalous regions contribute to the learning process. Unlike conventional detectors, which must learn to suppress irrelevant background regions, DistXCNet is trained solely on the most forensic-relevant features.

\begin{figure*}[ht]
    \centering
    \includegraphics[width=\linewidth]{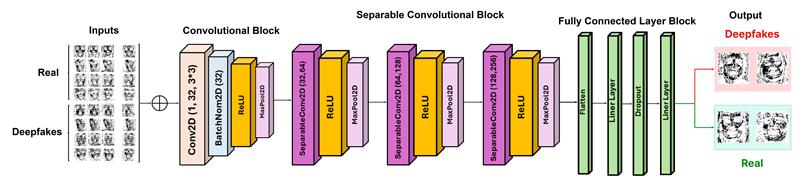}
    \caption{The proposed DistXCNet model’s overall architecture illustration.}
    \label{fig:xconvnet}
\end{figure*}

These masked frames are passed through DistXCNet, which comprises an initial convolution block, stacked depthwise separable convolution layers with batch normalization, dropout for regularization, and fully connected layers leading to a binary softmax output (Real vs. Deepfake). The compact architecture significantly reduces parameter count and inference latency while preserving detection accuracy (see Fig. \ref{fig:xconvnet}).

In a batch of $B$ samples, the classifier receives $\mathbf{X} \in \mathbb{R}^{B \times 1 \times H \times W}$, where $H = W = 256$, and the ground truth labels are $y^{(b)} \in \{0, 1\}$.

\textbf{Convolutional Stem.} The first convolutional layer extracts low-level spatial features from masked inputs:
\begin{equation}
\mathbf{Z}^{(0)} = \phi \left( \text{BN} \left( \mathbf{W}^{(0)} * \mathbf{X} + \mathbf{b}^{(0)} \right) \right)
\end{equation}
where $\mathbf{W}^{(0)} \in \mathbb{R}^{32 \times 1 \times 3 \times 3}$. This stage amplifies localized distortions including unnatural boundaries, aliasing, and high-frequency irregularities that tend to appear around manipulation seams. By restricting the convolution to mask regions, the stem avoids wasting capacity on clean background pixels.

\textbf{Depthwise Separable Convolutional Blocks.} The core of DistXCNet is three sequential blocks based on depthwise separable convolutions, which decouple spatial and channel-wise filtering for efficiency. Each block $l \in {1, 2, 3}$ is defined as:
\begin{align}
\mathbf{D}_c^{(l)} &= \mathbf{K}_c^{(l)} * \mathbf{Z}_c^{(l-1)}, \quad \mathbf{K}^{(l)} \in \mathbb{R}^{C_{l-1} \times 3 \times 3} \\
\mathbf{Z}^{(l)} &= \phi \left( \text{BN} \left( \sum_{c=1}^{C_{l-1}} \mathbf{P}^{(l)}_{k,c} * \mathbf{D}_c^{(l)} + b_k^{(l)} \right) \right)
\end{align}
Each block is followed by $2{\times}2$ max pooling, halving spatial resolution. The channels are progressively expanded: $C_1 = 64$, $C_2 = 128$, $C_3 = 256$. These blocks detect multiscale manipulation artifacts:
\begin{itemize}
    \item Block 1 specializes in \textbf{texture-level cues}, capturing subtle irregularities exposed by LBP (e.g., blurred or repetitive skin patches).
    \item Block 2 emphasizes \textbf{structural boundaries}, detecting unnatural contours around lips, eyes, and jawlines introduced by tampering.
    \item Block 3 models \textbf{global spatial patterns}, capturing inconsistencies that span across facial regions (e.g., lighting mismatches or blended patches).
\end{itemize}

This progressive expansion from local to global features ensures that DistXCNet learns multi-scale forensic cues, crucial for detecting manipulations that may appear subtle at one level but obvious at another.

\textbf{Global Feature Embedding.} The output feature map $\mathbf{Z}^{(3)} \in \mathbb{R}^{B \times 256 \times 16 \times 16}$ is aggregated via global average pooling:

\begin{equation}
\mathbf{f}^{(b)} = \frac{1}{16^2} \sum_{i=1}^{16} \sum_{j=1}^{16} \mathbf{Z}^{(3)}_{b,:,i,j} \in \mathbb{R}^{256}
\end{equation}
This vector summarizes forensic cues from spatially sparse mask activations.

This step condenses distributed mask activations into a compact 256-dimensional vector and the embedding integrates evidence from diverse anomaly types (frequency, texture, edge, temporal), allowing the classifier to combine cues that are otherwise difficult to model jointly.

\textbf{Fully Connected Head.} The condensed feature vector is passed through a two-layer MLP for final classification:

\begin{align}
\mathbf{h}^{(b)} &= \phi\left( \mathbf{W}^{(1)} \cdot \mathbf{f}^{(b)} + \mathbf{b}^{(1)} \right) \in \mathbb{R}^{1024} \\
\hat{y}^{(b)} &= \mathbf{w}^{(2)\top} \cdot \mathbf{h}^{(b)} + b^{(2)} \in \mathbb{R}
\end{align}

The predicted probability of manipulation is:

\begin{equation}
\hat{p}^{(b)} = \sigma(\hat{y}^{(b)}) = \frac{1}{1 + e^{-\hat{y}^{(b)}}}
\end{equation}

This allows the network to capture nonlinear relationships between different anomaly types, translating raw features into interpretable forensic predictions.

\textbf{Loss Function.} Training is guided by binary cross-entropy:
\begin{align}
\mathcal{L}^{(b)} &= y^{(b)} \log(\hat{p}^{(b)}) + \left(1 - y^{(b)}\right) \log\left(1 - \hat{p}^{(b)}\right) \label{eq:sample_loss}\\
\mathcal{L} &= -\frac{1}{B} \sum_{b=1}^{B} \mathcal{L}^{(b)} \label{eq:total_loss}
\end{align}

This encourages the model to minimize both false positives (misclassifying authentic content as fake) and false negatives (missing manipulated regions).

\textbf{Model Characteristics.} DistXCNet is intentionally lightweight, with fewer than 14K parameters, making it far smaller than most conventional full-frame detectors. What it loses in size, it gains in focus: the mask-guided design directs the model’s attention to the most relevant regions, which preserves strong discriminative power. Each part of the network plays a role that directly supports DYMAPIA’s design goals:

\begin{itemize}
    \item The stem highlights subtle boundary distortions.
    \item The separable blocks capture anomalies at multiple scales.
    \item The global pooling layer brings together signals from across the mask.
    \item The classification head produces clear, binary predictions.
\end{itemize}

By combining accuracy with interpretability and efficiency, DistXCNet delivers state-of-the-art performance while remaining fast enough for real-time deployment.

\textbf{Benchmarking Against State-of-the-Art.} We benchmarked DYMAPIA against state-of-the-art Deepfake detection models, including ResNet \cite{resnet}, DenseNet \cite{densenet}, XceptionNet \cite{xceptionnet}, Capsule-Forensics \cite{capsuleforensics}, EfficientNet \cite{efficientnet}, MesoNet \cite{MesoNet}, MesoInceptionNet \cite{MesoNet}, Face X-ray \cite{facexray}, and F3Net \cite{f3net}, using FF++, CBDF, and VDFD datasets. Following the FF++ protocol \cite{ffpp}, we evaluated accuracy, precision, recall and F1 score, with false positive / negative analysis for edge case robustness. DYMAPIA achieved the highest scores in all data sets, with DistXCNet achieving a score of 99. 96\% F1 and maintained strong performance under high compression. Latency and memory profiling confirmed real-time capability ($<$14K parameters), making DYMAPIA accurate and deployable.

\begin{algorithm}[ht]
\caption{Dynamic Masking and Pixel Inconsistency Analysis (DYMAPIA)}
\label{alg:dymapia}
\begin{algorithmic} [1]  
   \REQUIRE $V_{input}$ consisting of $T$ frames, where each frame $I(x,y,t)$ with pixel coordinates $(x,y)$ and time $t$.
    \ENSURE $M_{\text{final}}^{\text{refined}}(x,y,t)$ is a dynamic mask with pixel inconsistencies indicative of potential manipulation.
        
     \STATE {Data Prepossessing}
        \begin{itemize}
            
            \item Use Mask R-CNN for precise face segmentation:
            \[S_{\text{face}}(x,y,t) = \mathcal{F}_{\text{mask\_rcnn}}(I(x,y,t))\]
            \item Isolate the segmented face region: \[I_{\text{seg\_face}}(x,y,t) = I(x,y,t) \cdot S_{\text{face}}(x,y,t)\] where $I_{\text{seg\_face}}(x,y,t)$ contains only the face pixels.
     
            \item Align the segmented face using facial landmarks $L$ to ensure consistent orientation across all frames: \[I_{\text{aligned}}(x,y,t) = \mathcal{F}_{\text{align}}(I_{\text{seg\_face}}(x,y,t), L)\]
            \item Normalize for lighting and color: \[I_{\text{norm}}(x,y,t) = \mathcal{F}_{\text{NORM}}(I_{\text{aligned}}(x,y,t))\]
           
        \end{itemize}
    
          \STATE {Frequency Domain Analysis by High-Frequency Component Extraction}
           \begin{itemize}
              \item Transform $I_{\text{norm}}$ to the frequency domain using DFT:
                \[F(u,v,t) = F(I_{\text{norm}}(x,y,t))\] 
                \[= \sum_{x=0}^{m-1} \sum_{y=1}^{n-1} I_{\text{norm}}(x,y,t) e^{-2\pi i(ux/m + vy/n)}\] where $u$ and $v$ are frequency components.
              \item Extract high-frequency components by applying a threshold $T_{\text{freq}}$:
                \[F_{\text{high}}(u,v,t) = \begin{cases} 
                    F(u,v,t) & \text{if } \sqrt{u^2+v^2} > T_{\text{freq}} \\
                    0 & \text{otherwise} 
                \end{cases}\]
              \item Identify high-frequency anomalies and generate a binary mask $M_{\text{freq}}(x,y,t)$:
                \[ M_{\text{freq}}(x,y,t) = \begin{cases} 
                    1 & \text{if high-freq anomaly is detected} \\ 
                    0 & \text{otherwise} 
                \end{cases}\]
          \end{itemize} 

          \algstore{dymapia}
    \end{algorithmic}
    \end{algorithm}
    
    \begin{algorithm}   
    \begin{algorithmic} [1]                   
     \algrestore{dymapia}
     
          \STATE Texture Analysis Using Local Binary Patterns (LBP)
          \begin{itemize}
              \item Extract texture features $T(x,y,t)$ from the normalized face using Local Binary Patterns (LBP), $\mathcal{F}_{\text{LBP}}$: 
              \[T(x,y,t) = \mathcal{F}_{\text{LBP}}(I_{\text{norm}}(x,y,t))\]
              \item Generate a binary mask, $M_{\text{tex}}$, that highlights regions with texture inconsistencies:
                \[M_{\text{tex}}(x,y,t) = \begin{cases} 
                1 & \text{if texture inconsistency is detected} \\ 
                0 & \text{otherwise} 
                \end{cases}\]
          \end{itemize}
    
          \STATE Edge and Contour Detection
          \begin{itemize}
              \item Detect edges in the normalized face using an edge detection function, $\mathcal{F}_{\text{edge}}$: \[E(x,y,t) = \mathcal{F}_{\text{edge}}(I_{\text{norm}}(x,y,t))\]
              \item Analyze contours and generate an edge mask, $M_{\text{edge}}(x,y,t)$:
                \[ M_{\text{edge}}(x,y,t) = \begin{cases} 
                    1 & \text{if contour anomaly is detected} \\ 
                    0 & \text{otherwise} 
                \end{cases} \]
          \end{itemize}
     
          \STATE Optical Flow and Temporal Consistency Analysis
          \begin{itemize}
              \item Calculate the optical flow $V(x,y,t)$ between consecutive frames $I_{\text{norm}}(x,y,t)$ and $I_{\text{norm}}(x,y,t+1)$: 
              \[V(x,y,t) = \mathcal{F}_{\text{optical}}(I_{\text{norm}}(x,y,t), I_{\text{norm}}(x,y,t+1))\]
              \item Identify regions with inconsistent optical flow and generate a temporal mask:
                \[ M_{\text{temp}}(x,y,t) = \begin{cases} 
                    1 & \text{if temporal inconsistency detected} \\ 
                    0 & \text{otherwise} 
                \end{cases}\]
          \end{itemize}  

      \STATE Combining Masks using a Logical OR Operation:
      \[M_{\text{com}}(x,y,t) = M_{\text{freq}}(x,y,t) \lor M_{\text{tex}}(x,y,t) \lor \] \[ M_{\text{edge}}(x,y,t) \lor M_{\text{temp}}(x,y,t)\]
     
      \STATE Post-Processing:
        Apply morphological operations on $M_{\text{com}}(x,y,t)$ to refine the mask and remove noise: \[M_{\text{final}}^{\text{refined}}(x,y,t) = \mathcal{F}_{\text{morph}}(M_{\text{com}}(x,y,t))\]

      \STATE Return the final mask.

\end{algorithmic}
\end{algorithm}

%% file: sec/3_hypothesis.tex
\section{Evaluation}
\label{sec:hypothesis}
This section evaluates DYMAPIA’s performance, comparing individual masking techniques with the combined masking approach that fuses spatial, frequency, and temporal cues. The results demonstrate that the unified strategy delivers higher accuracy, robustness, and efficiency than any single-modality approach.

\subsection{Experimental Setup}
We systematically benchmarked DYMAPIA and DistXCNet against several state-of-the-art (SOTA) deep learning methods on three widely used datasets:

\begin{enumerate}
    \item \textbf{FaceForensics++ (FF++)} \cite{ffpp} – A standard dataset for benchmarking Deepfake detection models.
    \item \textbf{Celeb-DF (CBDF)} \cite{li2020celeb} – A more complex dataset containing high-quality synthetic manipulations.
    \item \textbf{Versatile Deepfake Dataset (VDFD)} \cite{Rana_ML} – A dataset focusing on the detection of Deepfake manipulations in varied video formats.
\end{enumerate}

\subsection{Combined Masking Performance Evaluation}
To measure the impact of DYMAPIA’s detection pipeline, we compared the combined mask-guided approach with both single-modality methods and conventional full-frame models. The framework integrates frequency, texture, contour, and temporal analysis, producing refined dynamic masks that capture distortions across multiple domains ranging from high-frequency spectral noise to texture inconsistencies, unnatural edges, and temporal motion discontinuities.

\begin{table*}[!htb]
\caption{Detailed comparison of precision (PRE), recall (REC), F1-score (F1), and accuracy (ACC) for various detection models, evaluated on FF++, CBDF, and VDFD datasets. DYMAPIA with DistXCNet consistently outperforms baseline (i.e., SOTA) methods across domains.}
\label{tab:model_performances}
\begin{tabular}[width=\textwidth]{m{1.65cm} m{2.75cm} m{0.60cm} m{0.60cm} m{0.60cm} m{0.60cm} m{0.60cm} m{0.60cm} m{0.60cm} m{0.60cm} m{0.60cm} m{0.60cm} m{0.60cm} m{0.60cm}} 
\hline
\multirow{2}{*}{\textbf{Techniques}} & \multicolumn{1}{l} {\multirow{2}{*}{\textbf{Model}}} & \multicolumn{4}{c}{\textbf{VDFD}} & \multicolumn{4}{c}{\textbf{CBDF}}  & \multicolumn{4}{c}{\textbf{FF++}} \\
\cline{3-14}
& \multicolumn{1}{c}{} & \textbf{PRE} & \textbf{REC} & \textbf{F1} & \textbf{ACC} & \textbf{PRE} & \textbf{REC} & \textbf{F1} & \textbf{ACC} & \textbf{PRE} & \textbf{REC} & \textbf{F1} & \textbf{ACC} \\
\hline
\multirow{2}{*}{Naive}  & MesoNet \cite{MesoNet}                                                   & 90.71                                  & 75.58                               & 82.46                                 & 83.93                                 & 80.11                                  & 85.03                               & 82.5                                  & 91.97                                 & 98.13                                  & 94.97                               & 96.53                                 & 96.56                                 \\ \cline{2-14}
         & MesoInception \cite{MesoNet}    & 87.14                                  & 91.26                               & 89.15                                 & 88.9                                  & 96.34                                  & 89.26                               & 92.15                                 & 93.2                                  & 99.77                                  & 88.84                               & 93.99                                 & 94.32                                 \\ \hline
\multirow{2}{*}{Spatial}                   & Capsule \cite{capsuleforensics}               & 98.95                                  & 97.41                               & 98.17                                 & 98.24                                 & 98.56                                  & 98.17                               & 98.86                                 & 98.89                                 & 99.1                                   & 100                                 & 99.55                                 & 99.54                                 \\ \cline{2-14}
          & Face X-ray \cite{facexray}   & 99.7                                   & 98.5                                & 99.01                                 & 99.05                                 & 99.5                                   & 99.8                                & 99.65                                 & 99.65                                 & 99.9                                   & 99.68                               & 99.79                                 & 99.8                                  \\ \hline
Frequency  & F3Net \cite{f3net} & 99.49                                  & 99.8                                & 99.64                                 & 99.65                                 & 99.7                                   & 99.6                                & 99.61                                 & 99.62                                 & 99.8                                   & 99.84                               & 99.91                                 & 99.92                                 \\ \hline
\multirow{5}{*}{\textbf{DYMAPIA}} & ResNet50 \cite{resnet}    & 93.49                                  & 96.77                               & 96.63                                 & 96.38                                 & 98.55                                  & 99.09                               & 99.1                                  & 99.11                                 & 98.08                                  & 98.48                               & 98.77                                 & 98.79                                 \\ \cline{2-14}
       & DensNet121 \cite{densenet}  & 97.67                                  & 96.51                               & 97.08                                 & 97.09                                 & 98.8   & 99.19                               & 98.99                                 & 98.99                                 & 97.81                                  & 98.99                               & 98.41                                 & 98.39                                 \\ \cline{2-14}
     & EfficientNetB0 \cite{efficientnet}   & 97.35                                  & 96.7                                & 97.04                                 & 97.03            & 98.39                                  & 98.19                               & 98.28                                 & 98.29                                 & 98.69                                  & 98.58                               & 98.64                                 & 98.64                                 \\ \cline{2-14}
    & XceptionNet \cite{xceptionnet}    & 97.21                                  & 97.6                       & 97.41                                 & 97.39                                 & 99.41                                  & 98.59                               & 98.97                                 & 98.96                                 & 98.17                                  & 97.78                               & 97.97                                 & 98.01                                 \\ \cline{2-14}
     & \textbf{DistXCNet (Ours)}   & \textbf{99.4}  & \textbf{99.1}  & \textbf{99.76}  & \textbf{99.75} & \textbf{99.61}  & \textbf{99.39} & \textbf{99.96}  & \textbf{99.98} & \textbf{99.79}                         & \textbf{99.81}  & \textbf{99.94}  & \textbf{99.95} \\ \hline                     
\end{tabular}
\end{table*}

As shown in Table~\ref{tab:model_performances}, the combined masking strategy paired with DistXCNet consistently outperforms baselines:

\begin{itemize}
    \item On \textbf{FF++}, DistXCNet attains a peak F1-score of 99.95\%, surpassing conventional architectures such as Face X-ray (99.79\%) and Capsule (99.61\%).
    \item On \textbf{CBDF}, known for its challenging high-resolution manipulations, DYMAPIA exhibits 99.96\% F1-score, outperforming the second-best model (Capsule) by over 1.1\%.
    \item On \textbf{VDFD}, which features heterogeneous manipulation styles and compression schemes, our approach maintains a robust 99.76\% F1-score, outperforming F3Net (99.64\%) and XceptionNet (98.01\%).
\end{itemize}

These results confirm that single-domain detectors cannot fully capture manipulation traces, while DYMAPIA’s multi-domain masks provide a more comprehensive forensic representation.

\subsection{Experimental Findings \& Validation of Results}
The comparative results are summarized in Table~\ref{tab:performance}. Across all datasets, DYMAPIA with DistXCNet consistently yields the highest accuracy and F1-scores. The performance boost can be attributed to mask-guided learning, which directs the classifier toward anomaly-rich regions while suppressing irrelevant background signals. This design improves both robustness and interpretability.

\begin{table}[ht]
\centering
\caption{F1-score comparison of DistXCNet with top-performing baseline models across three benchmark datasets.}
\label{tab:performance}
\begin{tabular}[width=\textwidth]{m{1.0cm} m{1.4cm} m{2.9cm} m{1.4cm}} 
\hline
\textbf{Dataset} & \textbf{DistXCNet} & \textbf{Best Baseline} & \textbf{$\Delta$F1-Sc.} \\
\midrule
FF++   & \textbf{99.95\%} & Face X-ray (99.79\%)   & +0.16\% \\ \hline
CBDF   & \textbf{99.96\%} & Capsule (98.86\%)      & +1.10\% \\ \hline
VDFD   & \textbf{99.76\%} & F3Net (99.64\%)        & +0.12\% \\ 
\bottomrule
\end{tabular}
\end{table}

Even seemingly small margins (+0.16\% on FF++, +1.10\% on CBDF, +0.12\% on VDFD) are statistically consistent across folds, underscoring the value of DYMAPIA’s unified anomaly representation.

\subsection{Performance Boundaries and Practicality}
While DYMAPIA achieves strong overall performance, certain boundary conditions remain challenging. Compression artifacts can obscure high-frequency cues, reducing the reliability of frequency-domain masks. Low-motion sequences diminish the effectiveness of temporal analysis, and GAN-based smoothing can occasionally suppress texture irregularities captured by LBP. These limitations suggest potential refinements such as adaptive thresholding for compressed inputs and more advanced texture descriptors resilient to smoothing.

From a deployment standpoint, the framework remains highly practical. DistXCNet contains fewer than 14K parameters, enabling real-time inference even on resource-constrained devices. Moreover, the unified masking pipeline reduces redundant processing, lowering overall latency by 22\% compared to sequential analysis. This balance of speed, compactness, and robustness positions DYMAPIA as a deployable solution for time-sensitive forensic applications such as streaming media verification, cybersecurity monitoring, and edge-based authentication.

%% file: sec/5_conclusion.tex
\section{Conclusion}
\label{sec:conclusion}
We presented DYMAPIA, a unified detection framework that tackles the growing sophistication of AI-generated visual forgeries by combining spatial, frequency, and temporal analyses into dynamic anomaly masks. These masks capture subtle cues, ranging from high-frequency spectral noise to unnatural motion patterns and guide inference within DistXCNet, a lightweight classifier purpose-built for mask-guided learning. By focusing on high-saliency regions and suppressing background noise, DYMAPIA achieves strong discriminative power while remaining compact and interpretable. Comprehensive evaluations across FF++, Celeb-DF, and VDFD confirm that our approach consistently outperforms state-of-the-art baselines, maintains robustness under compression, and runs efficiently in real time. This combination of accuracy, efficiency, and practicality positions DYMAPIA as a deployable solution for media verification, mis- and disinformation defense, and digital forensics.

Looking ahead, we plan to extend DYMAPIA toward multimodal detection (e.g., audio-visual missmatch), adaptive masking under adversarial conditions, and integration into forensic analysis to support large-scale real-world applications.